\documentclass[11pt,a4paper]{article}

\usepackage[utf8]{inputenc}
\usepackage[T1]{fontenc}
\usepackage{lmodern}
\usepackage[margin=2.5cm]{geometry}
\usepackage{setspace}
\usepackage{amsmath,amssymb,mathtools}
\usepackage{graphicx}
\usepackage{booktabs}
\usepackage{multirow}
\usepackage{xcolor}
\usepackage[authoryear]{natbib}
\makeatletter
\let\citep@natbib\citep
\def\citep#1{\ifhmode\unskip\space\fi\citep@natbib{#1}}
\makeatother
\usepackage[labelfont=bf,labelsep=period,font=small]{caption}
\usepackage{authblk}
\usepackage{microtype}
\usepackage[hidelinks]{hyperref}

\graphicspath{{./}}
\makeatletter
\renewcommand\@seccntformat[1]{\csname the#1\endcsname
  \ifnum\pdfstrcmp{#1}{section}=0 .\fi\quad}
\makeatother
\newcommand{\figref}[1]{Fig.~\ref{#1}}

\title{\textbf{Neural strategy of cognitive map representation informs
position encoding in visual transformer}}

\author[1,$\dagger$]{Tianhao Chu}
\author[1,2,3,$\dagger$]{Sichao He}
\author[1,2,3,$\dagger$]{Yuling Wu}
\author[1,$\dagger$]{Shutian Liang}
\author[1,2,3,$*$]{Si Wu}

\affil[1]{School of Psychological and Cognitive Sciences, Peking University, China.}
\affil[2]{Peking-Tsinghua Center for Life Sciences, Academy for Advanced Interdisciplinary Studies, IDG/McGovern Institute for Brain Research, Key Laboratory of Machine Perception (Ministry of Education), Peking University, China.}
\affil[3]{Center of Quantitative Biology, Peking University, China.\protect\\[2pt]
$^{\dagger}$These authors contributed equally to this work.\protect\\
$^{*}$Correspondence: \texttt{siwu@pku.edu.cn}}
\date{}

\begin{document}

\maketitle

\begingroup
\renewcommand{\thefootnote}{}
\footnotetext{Abbreviations: AI, artificial intelligence; CV, coefficient of
variation; GL-PE, Grid-Learned Positional Encoding; Hex-RoPE, hexagonal rotary
position embedding; HSR, Hexagonal Spiral RoPE; LPE, learned positional
embedding; MEC, medial entorhinal cortex; nRMSE, normalized root-mean-square
error; PE, positional embedding; RNN, recurrent neural network; RoPE, rotary
position embedding; ViT, vision transformer; WEF, Weierstrass elliptic
function.}
\endgroup

\begin{abstract}
\noindent
\textbf{Background.} Band cells and grid cells coexist in the cognitive map of
both biological navigation circuits and optimization-based artificial
intelligence (AI) models, yet their distinct computational functions and
division of labour in spatial cognition remain unclear.

\textbf{Methods.} We combine numerical circuit simulations, theoretical analysis
and neural network optimization to dissect this division of labour and to
carry the resulting spatial metric into transformer position encoding.

\textbf{Results.} Band cells perform neuron-efficient path integration, whereas
grid cells form an isotropic metric representation of two-dimensional
Euclidean space. Projecting two-dimensional velocity onto one-dimensional band
attractors avoids the quadratic neuronal cost of tiling a two-dimensional
manifold, whereas preserving isotropic Euclidean ($\ell_2$) rather than
axis-aligned Manhattan ($\ell_1$) distances between locations requires the
hexagonal symmetry of grid cells. Viewed from this cognitive-map perspective,
standard positional encodings in visual transformers adopt a band-cell-like
axial factorization that distorts Euclidean distances. Introducing grid-cell
Euclidean metric representation into positional encoding reduces directional
anisotropy more than eightfold and consistently improves recognition across
five visual tasks.

\textbf{Conclusions.} These results distill a general neural encoding principle
for spatial cognition that directly advances AI vision models.
\end{abstract}

\section{Introduction}

Constructing an internal cognitive map of external space is a
foundational computation of biological and artificial spatial
cognition\citep{mcnaughton2006path}. Across the mammalian medial
entorhinal cortex (MEC) and adjacent parasubiculum, the neural
representation features two distinct spatially periodic cell types: grid
cells fire at the vertices of a two-dimensional hexagonal lattice
organized into discrete scale
modules\citep{hafting2005microstructure,stensola2012entorhinal}, whereas
band cells fire along parallel stripes periodic along a single preferred
orientation\citep{krupic2012band}. This joint band- and grid-cell
representation appears not only in biological navigation circuits but
also in optimization-based AI models---specifically recurrent neural
networks (RNNs) trained to path-integrate velocity
trajectories\citep{fuhs2006spin,burgess2007oscillatory,burak2009accurate,bush2015using,cueva2018emergence,banino2018vector,sorscher2023unified,chu2025unfolding}.
Our recent circuit dissection of task-trained RNNs showed that upstream
band-cell modules receive velocity inputs and project to downstream
grid-cell populations for position
readout\citep{chu2025unfolding}, while normative coding theories have
examined how hexagonal grid lattices alone embed Euclidean
space\citep{fiete2008what,wei2015principle,stemmler2015connecting,dordek2016extracting,gao2019learning,xu2025conformal}.
However, what distinct computational functions band cells and grid cells
perform within the cognitive map---and why spatial cognition requires a
division of labour between them---remains unresolved.

Here we combine numerical simulations of continuous-attractor circuits,
theoretical analysis of population geometry, and optimization of deep
metric-learning models to establish the division of labour between band
and grid cells. Band cells are specialized for economical path
integration, grid cells for the isotropic metric representation of
two-dimensional Euclidean space. Tiling a two-dimensional continuous
attractor requires a neuron count that scales quadratically with linear
resolution, so factorizing velocity integration into one-dimensional
ring attractors lets band cells integrate self-motion cues with
substantially fewer neurons than two-dimensional grid sheets alone. Yet a
useful cognitive map must also preserve geometric distance relationships
between locations, not merely discriminate them. Optimizing a neural
network without velocity inputs to preserve local inter-location
distances, and conversely analysing mathematically constructed reference
populations, shows that the spatial metric directly selects the cell
type: band-cell codes naturally track axis-aligned Manhattan ($\ell_1$)
distance, whereas preserving isotropic Euclidean ($\ell_2$) distance
drives the emergence of hexagonal grid cells.

This cognitive-map perspective provides a new lens on how contemporary AI
visual transformers encode two-dimensional spatial
positions\citep{vaswani2017attention,dosovitskiy2020image,liu2021swin,chu2021conditional}.
A visual transformer processes static image patches without velocity
integration and thus needs a purely representational cognitive map of the
image plane, yet we find that standard two-dimensional positional
encodings---most notably axial rotary position embedding
(2D RoPE)\citep{su2021roformer,heo2024rotary}---adopt a band-cell-like
orthogonal channel factorization that tracks Manhattan-like axial
structure and distorts Euclidean distances between patches. We therefore
introduce the grid-cell principle of Euclidean metric representation
into visual positional encoding through two complementary algorithms:
Grid-Learned Positional Encoding (GL-PE), which regularizes learnable
positional embeddings toward local Euclidean isometry, and Hexagonal
Spiral RoPE (HSR), which rotates three $120^\circ$-separated hexagonal
axes progressively across attention heads. This restores Euclidean
geometry across the image plane, reducing directional kernel anisotropy
more than eightfold and consistently improving recognition across five
visual datasets---natural objects, satellite land-use imagery, retinal
scans and biometric identification---and distilling a general encoding
principle for spatial cognition with direct utility for AI vision
architectures.

\section{Results}

\subsection{Division of labour between band cells and grid cells in cognitive map representation}

To determine why band cells and grid cells co-emerge in biological and
artificial cognitive maps, we dissected their contributions to the two
core computations of spatial cognition: dynamical path integration of
self-motion cues and static metric representation of physical space
(\figref{fig:coexistence}a--d).

\subsubsection{Band cells enable neuron-efficient path integration}

\begin{figure}[!htbp]
  \centering
  \includegraphics[width=\linewidth]{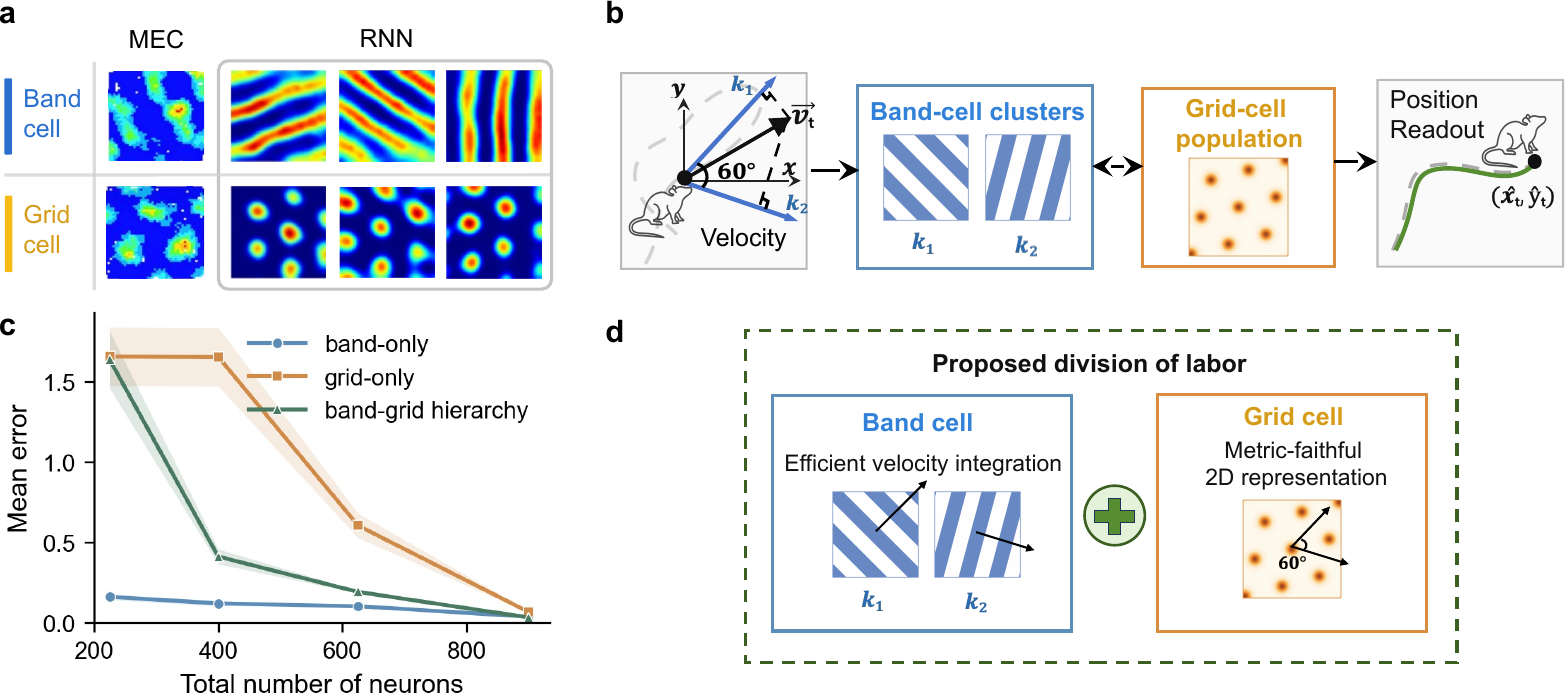}
  \caption{\textbf{Division of labour between band cells and grid cells
  in cognitive map representation.}
  \textbf{a}, Spatially periodic band-like and hexagonal grid-like firing
  patterns in the medial entorhinal cortex (MEC) and in recurrent neural
  networks (RNNs) trained for path integration; representative maps adapted
  from \citet{krupic2012band} and \citet{chu2025unfolding}, respectively.
  \textbf{b}, Hierarchical information flow in task-trained path-integration
  networks\citep{chu2025unfolding}: two-dimensional velocity $\mathbf{v}_t$
  projects onto directional band-cell clusters, which interact recurrently
  with a grid-cell population supporting position readout
  $(\hat{x}_t,\hat{y}_t)$.
  \textbf{c}, Mean Euclidean position decoding error over 1,200-step
  trajectories for band-only, grid-only and band--grid circuits at matched
  neuronal budgets; curves are means over 30 paired seeds and shading
  indicates 95\% confidence intervals (truncated at zero).
  \textbf{d}, Proposed division of labour: band cells integrate velocity in
  one dimension at neuron-efficient cost, whereas grid cells form a
  metric-faithful two-dimensional Euclidean representation.}
  \label{fig:coexistence}
\end{figure}

To examine how band-cell and grid-cell circuit architectures differ in
neuronal efficiency for path integration, we first performed numerical
simulations comparing three continuous-attractor models in a
path-integration task (\figref{fig:coexistence}a,b): a band-only circuit of
one-dimensional ring attractors, a grid-only circuit of two-dimensional
toroidal attractors, and the composite band--grid hierarchy identified in
task-trained RNNs\citep{chu2025unfolding}. Each circuit integrated
identical noisy velocity trajectories across matched total budgets of
225, 400, 625 and 900 dynamical neurons, and we evaluated path-integration
accuracy by decoding two-dimensional position from population activity
along each trajectory (Methods).

Across all restricted neuron budgets, the band-only circuit achieved
substantially lower position decoding error than the grid-only circuit
(\figref{fig:coexistence}c). The band-only circuit already maintained
accurate path integration at the smallest budget of 225 neurons and
improved steadily as the budget increased. By contrast, at 225 and 400
neurons the grid-only circuit suffered roughly tenfold larger errors and
required 900 neurons to approach the precision the band-only circuit
achieved with 225 neurons, while the band--grid hierarchy reached
intermediate error once its downstream grid sheet had sufficient
resolution, from 400 neurons onward. The gap arises because band
modules factorize two-dimensional velocity integration into
one-dimensional ring attractors, whereas a grid-only circuit must tile a
two-dimensional torus with enough neurons for stable, drift-free
activity bumps. One-dimensional band modules therefore suffice for
accurate, neuron-efficient path integration, indicating that
two-dimensional grid cells contribute to the cognitive map through
spatial representation rather than velocity integration itself
(\figref{fig:coexistence}d).

\subsubsection{Grid cells provide an isotropic Euclidean metric representation of space}

\begin{figure}[!htbp]
  \centering
  \includegraphics[width=0.99\linewidth]{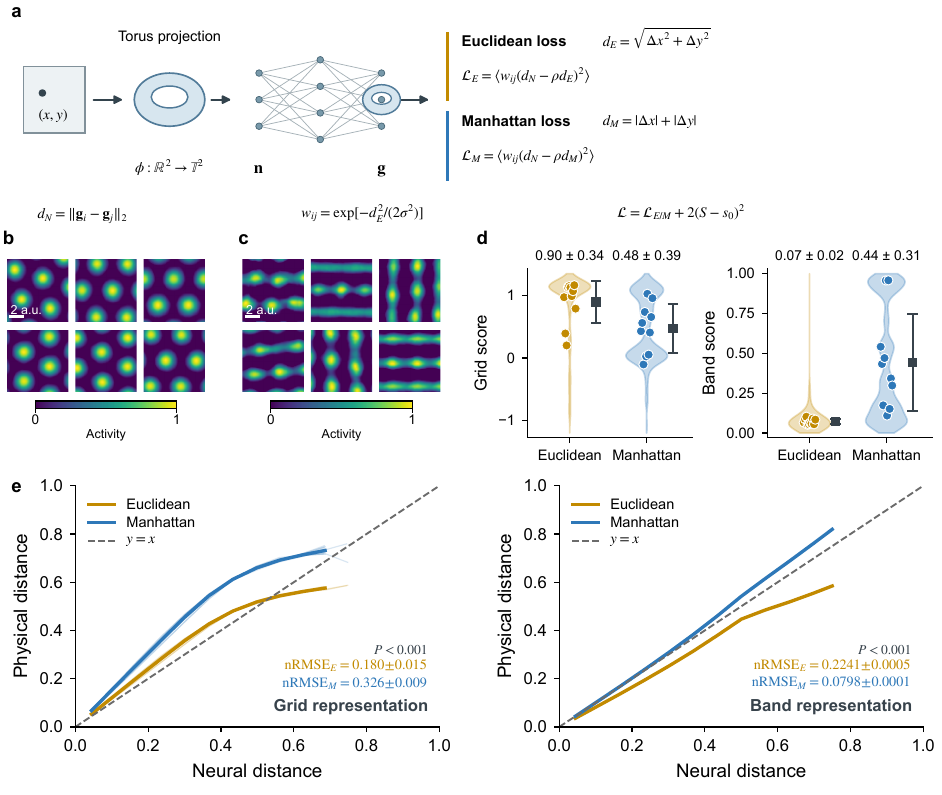}
  \caption{\textbf{The target spatial metric selects between grid-cell
  and band-cell representations.}
  \textbf{a}, Feedforward metric-learning architecture: static coordinates
  $(x,y)$ map to a 32-unit non-negative, unit-norm response trained to
  match either Euclidean ($d_E$, yellow) or Manhattan ($d_M$, blue)
  distances within a shared Gaussian neighbourhood envelope
  ($w_{ij}=\exp[-d_E^2/(2\sigma^2)]$), with isometry weight 1 and
  population-dispersion regularizer weight 2.
  \textbf{b},\textbf{c}, Six representative firing-rate maps from
  Euclidean-trained (\textbf{b}) and Manhattan-trained (\textbf{c})
  networks, selected by the same condition-blind score-central rule with
  at most two units per network; maps share spatial extent, receive no
  smoothing or rotation, and are peak-normalized to $[0,1]$.
  White scale bars denote 2 arena units.
  \textbf{d}, Continuous grid- and band-score distributions. Dots are
  individual network means ($n=10$ paired seeds per condition); squares,
  error bars and headers give the across-network mean $\pm$ s.d. Violin
  contours show the network-weighted density of unit scores with
  well-defined scores (grid: 117/320 Euclidean, 153/320 Manhattan; band:
  118/320 and 153/320; Supplementary Note~1).
  \textbf{e}, Local distance-preservation curves for ten independently
  constructed 32-unit analytic populations per representation: a
  three-axis hexagonal grid code (left) and two orthogonal 16-unit band
  modules (right). Physical Euclidean and Manhattan distances are plotted
  against neural distance within $d_E\le0.6$ under one
  representation-level, metric-neutral scale factor (Methods). Faint
  curves show individual seeds; thick curves and shading are mean
  $\pm$ s.d.\ across seeds (grid $\mathrm{nRMSE}$: $0.180\pm0.015$
  Euclidean versus $0.326\pm0.009$ Manhattan; band $\mathrm{nRMSE}$:
  $0.0798\pm0.0001$ Manhattan versus $0.2241\pm0.0005$ Euclidean; both
  paired $P<0.001$); dashed lines indicate $y=x$.}
  \label{fig:metric}
\end{figure}

Spatial localization involves two distinct
computational steps: integrating velocity cues over time, and forming a
representation of spatial location. Merely distinguishing individual
positions is not sufficient; an effective spatial representation must
also capture the pairwise distances between positions, which point-wise
decoding accuracy cannot evaluate. To examine how preserving spatial
distances shapes the cognitive map independently of velocity integration,
we optimized a feedforward network mapping static coordinates
$\mathbf{p}=(x,y)^{\top}$ to population responses
$\mathbf{z}_{\mathbf{p}}$ whose pairwise neural distances
$\|\mathbf{z}_{\mathbf{p}}-\mathbf{z}_{\mathbf{q}}\|_2$ match local
physical distances (\figref{fig:metric}a; Methods) under either a
Euclidean ($\ell_2$) or Manhattan ($\ell_1$) target metric:
\begin{equation}
  d_E(\mathbf{p},\mathbf{q})=\|\mathbf{p}-\mathbf{q}\|_2,
  \qquad
  d_M(\mathbf{p},\mathbf{q})=\|\mathbf{p}-\mathbf{q}\|_1 .
  \label{eq:metrics}
\end{equation}

Under otherwise identical optimization settings (Methods), changing the
target metric alone switched the emergent cell types of the learned
population code (\figref{fig:metric}b--d). Networks optimized to
preserve local Euclidean distance consistently developed grid cells with
hexagonally arranged firing fields (\figref{fig:metric}b), whereas
those optimized to preserve Manhattan distance developed band cells
tuned to parallel periodic stripes (\figref{fig:metric}c). Under two
coordinate input schemes---a sinusoidal toroidal input layer
(\figref{fig:metric}b--d) or raw $(x,y)$ coordinates fed directly to the
multilayer perceptron (Supplementary Fig.~1 and Supplementary Note~1)---
metric selectivity was consistent: the Euclidean objective produced
significantly higher population gridness scores than the Manhattan
objective ($P=0.012$ with toroidal input; $P=0.0025$ with raw
coordinates), whereas the Manhattan objective produced significantly
higher bandness scores ($P=0.0047$ and $P=0.00037$,
respectively). Thus, while periodic boundary priors sharpen overall
lattice regularity, the target spatial metric itself determines whether
the cognitive map organizes into hexagonal grid cells or directional band
cells.

To conversely validate the intrinsic distance-encoding preferences of the
two cell types, we manually constructed idealized grid-cell and band-cell
populations and evaluated how well their neural distances preserve
Euclidean versus Manhattan distance (\figref{fig:metric}e; Methods).
A grid-cell code built from three wave vectors separated by $120^\circ$
distributes phase variations evenly across orientations, so local neural
distances remain isotropic and proportional to Euclidean distance
$\|\Delta\mathbf{p}\|_2$, whereas two orthogonal band-cell modules
decompose displacements independently along the $x$ and $y$ axes, so
their combined response tracks the axis-aligned contours of Manhattan
distance $\|\Delta\mathbf{p}\|_1=|\Delta x|+|\Delta y|$. Under a common
metric-neutral unit conversion calibrated on infinitesimal axis-aligned
displacements (Methods), the constructed hexagonal grid-cell populations
matched physical Euclidean distance with nearly half the normalized
root-mean-square error ($\mathrm{nRMSE}$) of Manhattan distance
($P<0.001$), whereas the orthogonal band-cell populations matched
Manhattan distance with nearly threefold lower error than Euclidean
distance ($P<0.001$; \figref{fig:metric}e), consistent with the trained
networks (Supplementary Fig.~2 and Supplementary Note~1).

Synthesizing our circuit simulations (\figref{fig:coexistence}c),
network optimization (\figref{fig:metric}b--d) and population analysis
(\figref{fig:metric}e) reveals the neural strategy underlying cognitive
map representation\citep{chu2025unfolding}. Band modules decompose
two-dimensional velocity $\mathbf{v}$ into one-dimensional scalars
$v_s=\mathbf{e}_s^{\top}\mathbf{v}$ integrable on ring attractors with
minimal neuronal overhead, but an orthogonal pair of band modules
inevitably distorts two-dimensional Euclidean distances at finite spatial
offsets. Grid cells resolve this trade-off by binding at least three
$120^\circ$-separated directional phases into a joint cognitive map whose
local metric stays isotropic and Euclidean.

\subsection{Cognitive-map-inspired position encoding for visual transformers}

Viewed through this cognitive-map lens, a visual transformer in AI faces
the same core spatial problem as the entorhinal cortex: building an
internal map of two-dimensional image space. We therefore examined
existing positional encoding algorithms from this band--grid perspective
and translated the grid-cell Euclidean metric principle into new designs
(\figref{fig:pe} and \figref{fig:tasks}).

\subsubsection{Grid-cell metric representation guides isotropic position encoding}

\begin{figure}[!htbp]
  \centering
  \includegraphics[width=\linewidth]{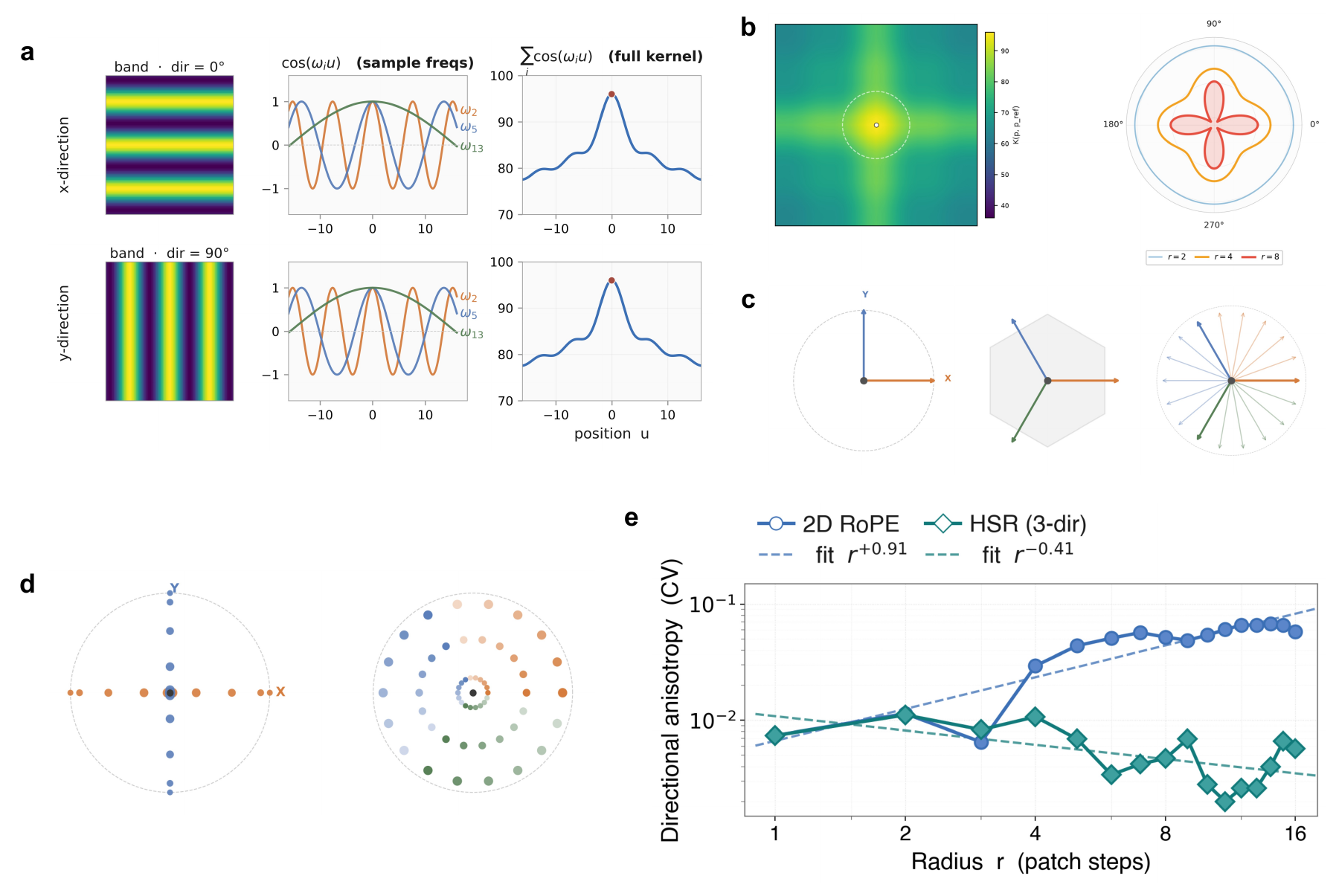}
  \caption{\textbf{Grid-cell metric representation guides the design of
  isotropic position encoding for visual transformers.}
  \textbf{a}, Structural correspondence between orthogonal band cells
  and axial 2D positional encodings: stripes along $x$ and $y$ (left),
  multi-frequency cosine components (middle) and similarity profiles
  (right).
  \textbf{b}, Analytic positional similarity field
  $K(\mathbf{p},\mathbf{q}_{\mathrm{ref}})$ of axial 2D RoPE around the
  origin reference position (left; dashed circle marks $r=8$ patch steps)
  and its polar angular profiles at radii $r=2$, 4 and 8 (right),
  showing fourfold cross-shaped concentration along the coordinate axes.
  \textbf{c}, Geometric progression from two orthogonal band-like axes
  (left) to three $120^\circ$-separated hexagonal axes within a module
  (middle) to hexagonal triads rotated progressively across heads
  (right).
  \textbf{d}, Two-dimensional Fourier frequency support of axial 2D RoPE,
  restricted to the coordinate axes (left), versus three-direction HSR
  with head-wise spiral offsets $\beta_h=2\pi h/(3H)$ (right).
  \textbf{e}, Directional anisotropy as the coefficient of variation
  $\mathrm{CV}(r)$ of the analytic similarity kernel over 64 angles on
  the $33\times33$ patch grid, versus radius $r\in\{1,\ldots,16\}$ patch
  steps under a matched per-head budget ($d_{\mathrm{head}}=192$, $H=6$,
  base $10{,}000$; log--log axes with power-law trend lines).}
  \label{fig:pe}
\end{figure}

Whereas biological navigation circuits must both integrate velocity and
represent space, a visual transformer operates purely in the static
representational regime of a cognitive map: image patches arrive with
known coordinates on a two-dimensional grid and require no temporal
velocity integration\citep{dosovitskiy2020image}. Just as in biological
spatial cognition, however, self-attention requires positional encodings
not merely to distinguish individual patch tokens, but to represent the
Euclidean distance and directional relationships between patches across
the image plane---a two-dimensional Euclidean metric structure that
existing positional encodings fail to provide. Unconstrained learned
positional embeddings (LPE) assign an independent trainable vector
$\mathbf{e}_{\mathbf{p}}$ to each patch position
$\mathbf{p}=(x,y)^{\top}$ without enforcing local inter-patch distance
relationships, whereas standard axial two-dimensional rotary position
embedding (2D RoPE) adopts a band-cell-like scheme: it splits
the query and key channels of each attention head into two orthogonal
halves rotated independently along $x$ and
$y$\citep{su2021roformer,heo2024rotary}. Under this band-like axial
factorization, the positional contribution to the attention score
between patches decomposes additively along the cardinal axes
(\figref{fig:pe}a):
\begin{equation}
  a_{\mathrm{axial}}(\Delta x,\Delta y)=F(\Delta x)+G(\Delta y).
\end{equation}
Because each channel group is invariant to displacements orthogonal to
its axis, axial 2D RoPE mirrors a pair of orthogonal band modules,
approximating two-dimensional proximity by summing two one-dimensional
axial kernels. Although nearly isotropic for immediately adjacent
patches, this band-cell-like encoding concentrates positional similarity
into a pronounced cross-shaped ridge along the horizontal and vertical
axes at finite patch separations, making it unfavourable for representing
Euclidean distances across the image plane (\figref{fig:pe}b).

To introduce the core principle of grid-cell Euclidean distance
representation into visual transformers (\figref{fig:pe}c), we designed
two complementary positional encoding algorithms (Methods). First,
within the additive learned family, Grid-Learned Positional Encoding
(GL-PE) regularizes the trainable positional embedding table
$\mathbf{e}_{\mathbf{p}}$ with a Gaussian-windowed local
Euclidean distance loss mirroring equation~\eqref{eq:metrics},
encouraging pairwise embedding distances to form a locally isometric
Euclidean map of the patch grid. Second, within the parameter-free
rotary family, Hexagonal Spiral RoPE (HSR) replaces the two orthogonal
band-like axes in each attention head with a grid-cell-like triad of
three unit wave directions $\{\mathbf{e}_1,\mathbf{e}_2,\mathbf{e}_3\}$
separated by $120^\circ$ (\figref{fig:pe}c,d). Because a single
three-axis triad samples only six directions in frequency space and
still leaves sixfold angular ripples at larger spatial offsets, head $h$
rotates its triad by a progressive offset angle
\begin{equation}
  \mathbf{e}_{s,h}=R(\beta_h)\,\mathbf{e}_s,\qquad \beta_h=\frac{2\pi h}{3H},
\end{equation}
so that the $H$ attention heads jointly tile orientation space uniformly
while preserving exact relative-position equivariance within each head
(Methods).

Evaluating the closed-form positional similarity kernel
$K(\mathbf{p},\mathbf{q}_{\mathrm{ref}})$ across spatial scales before
network training confirmed that HSR eliminates the directional
anisotropy of band-like encodings (\figref{fig:pe}e, Table~\ref{tab:cv}
and Methods). At a single patch step, all rotary encodings were nearly
isotropic on the discrete patch grid. As patch separation grew from
local to long-range distances, however, the directional coefficient of
variation ($\mathrm{CV}$) of band-like axial 2D RoPE climbed nearly
eightfold due to its fourfold axial ridges (\figref{fig:pe}b), and
unrotated Hex-RoPE (three $120^\circ$ axes with no cross-head offset,
$\beta_h=0$) climbed similarly due to coherent sixfold harmonics across
heads. By contrast, three-direction HSR remained isotropic across all
spatial scales, reducing mean directional distortion more than eightfold
relative to both axial 2D RoPE and unrotated Hex-RoPE on the discrete
patch grid, and 90-fold on continuous polar circles free of
pixel-lattice rounding (Table~\ref{tab:cv}). Ablating either design
component---reducing the three $120^\circ$ axes per head to two
$60^\circ$ axes, or removing the cross-head angular offsets
$\beta_h$---substantially degraded rotational isotropy
(Table~\ref{tab:cv}, lower block), showing that within-head
hexagonal balance and cross-head orientation tiling are jointly required
to preserve Euclidean isotropy at local and long-range patch separations.

\begin{table}[t]
  \centering
  \small
  \caption{\textbf{Directional distortion of rotary positional kernels.}
  Coefficient of variation $\mathrm{CV}(r)=\sigma(r)/|\mu(r)|$ of the
  analytic similarity kernel across 64 equispaced directions at
  representative radii (in patch steps) and averaged over $r=1$--16,
  on the discrete $33\times33$ patch grid (rightmost column: exact
  continuous polar-circle mean CV). Lower values indicate greater
  rotational isotropy. All encodings share an identical per-head
  dimensional and frequency budget ($d_{\mathrm{head}}=192$, $H=6$, base
  $10{,}000$). The lower block evaluates the $2\times2$ design crossing
  direction count per head (32 scales per direction for 3-dir versus 48
  for 2-dir under the 2D-RoPE-aligned frequency range; Methods) with the
  head-wise spiral offset $\beta_h=2\pi h/(3H)$.}
  \label{tab:cv}
  \setlength{\tabcolsep}{4.5pt}
  \begin{tabular}{lcccccc}
    \toprule
    Encoding & $r=1$ & $r=4$ & $r=8$ & $r=16$ & Mean ($r=1$--16) & Polar mean \\
    \midrule
    2D RoPE                      & 0.0074 & 0.0294 & 0.0519 & 0.0578 & 0.0465 & 0.0451 \\
    Hex-RoPE (3 axes, no offset) & 0.0105 & 0.0193 & 0.0631 & 0.0839 & 0.0500 & 0.0462 \\
    HSR (2-dir)                  & 0.0075 & 0.0242 & 0.0207 & 0.0250 & 0.0212 & 0.0199 \\
    HSR (3-dir)                  & 0.0074 & 0.0107 & 0.0047 & 0.0057 & \textbf{0.0056} & \textbf{0.0005} \\
    \midrule
    \multicolumn{7}{l}{\emph{Direction count $\times$ head-wise offset (fixed $d_{\mathrm{head}}=192$ budget)}} \\
    3 directions, with offset    & 0.0074 & 0.0107 & 0.0047 & 0.0057 & 0.0056 & 0.0005 \\
    2 directions, with offset    & 0.0075 & 0.0242 & 0.0207 & 0.0250 & 0.0212 & 0.0199 \\
    3 directions, no offset      & 0.0074 & 0.0116 & 0.0388 & 0.0432 & 0.0287 & 0.0264 \\
    2 directions, no offset      & 0.0107 & 0.0535 & 0.0639 & 0.0771 & 0.0596 & 0.0588 \\
    \bottomrule
  \end{tabular}
\end{table}

\subsubsection{Grid-inspired position encodings improve recognition across diverse visual tasks}

\begin{figure}[!htbp]
  \centering
  \includegraphics[width=\linewidth]{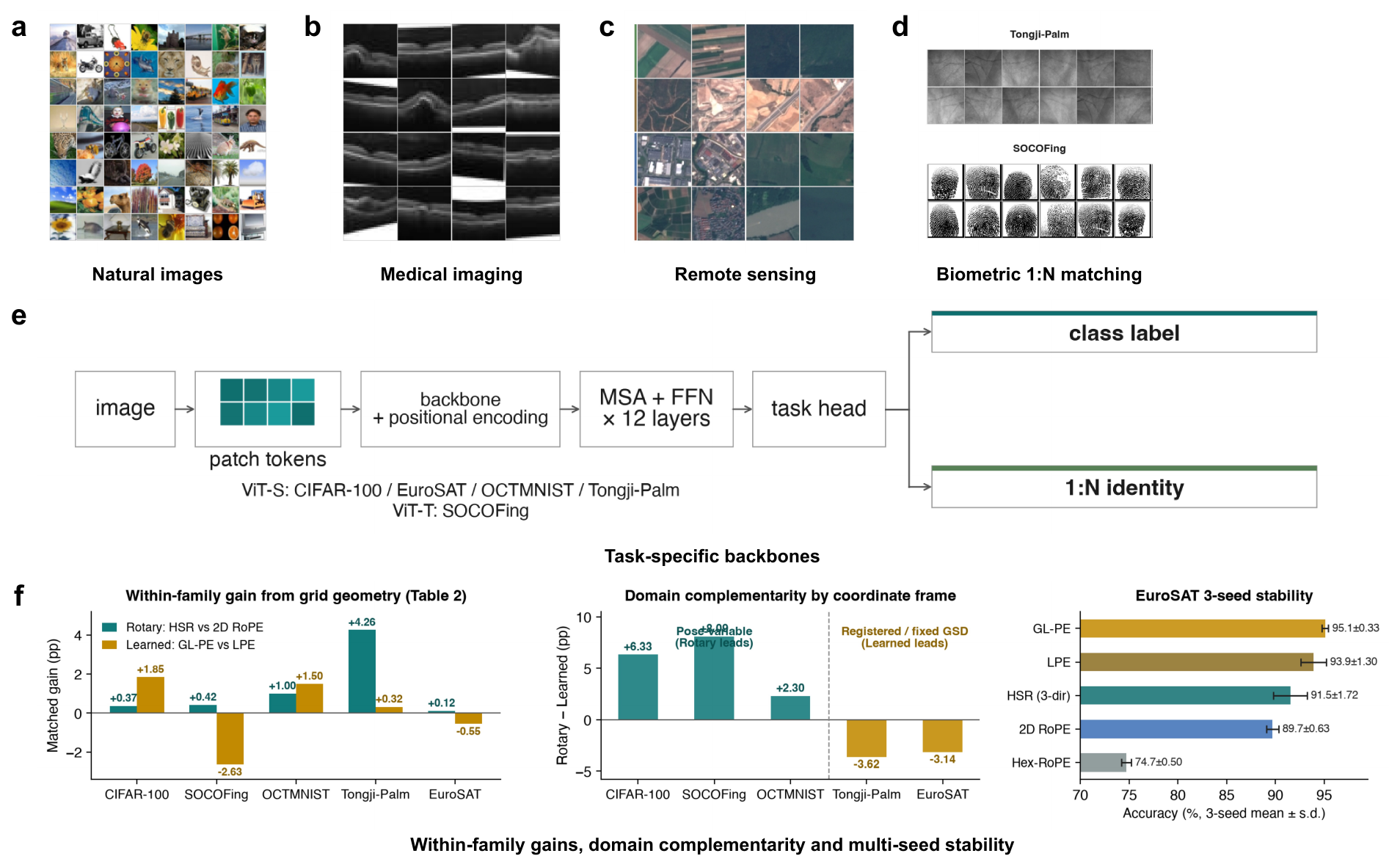}
  \caption{\textbf{The five evaluation datasets span distinct visual
  domains and spatial structures.}
  Representative training images at the input resolution supplied
  to the network. \textbf{a}, Natural objects (CIFAR-100, $32\times32$).
  \textbf{b}, Retinal optical coherence tomography B-scans (OCTMNIST,
  $28\times28$). \textbf{c}, Multispectral satellite land-use imagery
  (EuroSAT RGB, $224\times224$). \textbf{d}, Biometric $1:N$
  identification on palmprint ROIs (Tongji-Palm, $128\times128$) and
  fingerprints (SOCOFing, $128\times128$), where ridge and crease
  textures recur across individuals and identity relies on global
  two-dimensional geometry. \textbf{e}, Unified evaluation pipeline:
  input images are tokenized into non-overlapping patches, combined with
  the positional encoding under test, and processed by a shared 12-layer
  visual transformer backbone (ViT-Small for CIFAR-100, EuroSAT,
  OCTMNIST and Tongji-Palm; ViT-Tiny for SOCOFing) plus task head.
  \textbf{f}, Quantitative summary of Table~\ref{tab:accuracy}: (left)
  within-family matched gains of the best HSR variant over axial 2D
  RoPE (teal) and of GL-PE over unregularized LPE (amber); (middle)
  best-rotary minus best-learned accuracy, separating pose-variable
  domains (CIFAR-100, SOCOFing, OCTMNIST) from spatially registered
  domains (Tongji-Palm, EuroSAT); (right) EuroSAT three-seed mean
  $\pm$ s.d.\ (Supplementary Table~3).}
  \label{fig:tasks}
\end{figure}

We next tested whether endowing visual transformers with a grid-like
Euclidean cognitive map of image space improves recognition across five
datasets spanning distinct visual domains (\figref{fig:tasks}a--e and
Supplementary Tables~1 and 2): CIFAR-100\citep{krizhevsky2009learning},
OCTMNIST\citep{yang2023medmnist}, EuroSAT\citep{helber2019eurosat}, and
closed-set $1:N$ identification on SOCOFing\citep{shehu2018socofing} and
Tongji-Palm\citep{zhang2017towards}. On each dataset we compared GL-PE
against unconstrained LPE in the learned absolute family, and two- and
three-direction HSR against band-like axial 2D RoPE\citep{heo2024rotary},
unrotated Hex-RoPE and three recent non-axial encodings---multiscale
Grid-based PE\citep{li2024gridpe}, Weierstrass elliptic function (WEF)
features\citep{xin2025wefpe} and 16-direction plane-rotated
SpiralRoPE\citep{liu2026spiralrope}---under matched backbones and
training protocols (Methods).

Across all five visual tasks, our grid-inspired encodings consistently
outperformed the band-like axial 2D RoPE baseline and achieved the
highest accuracy on four of the five datasets
(Table~\ref{tab:accuracy} and Supplementary Table~4). Within the rotary
family, three-direction HSR ranked first among all nine encodings on
CIFAR-100 and SOCOFing fingerprint identification, while two-direction
HSR ranked first on OCTMNIST retinal scans. Comparing HSR against the
non-axial baselines showed why both within-head hexagonal balance and
cross-head orientation tiling are needed for downstream generalization.
Without cross-head offsets, unrotated Hex-RoPE lagged behind
three-direction HSR on four of the five tasks and dropped by 20
percentage points on EuroSAT, showing that sixfold lattice symmetry
aligned across heads can induce severe periodic aliasing. Conversely,
SpiralRoPE\citep{liu2026spiralrope} spreads each head's channel budget
across 16 rotation directions at the expense of frequency resolution per
direction, yielding accuracy drops of more than 12 percentage points
relative to three-direction HSR on both fine-grained biometric
identification tasks. Three-direction HSR resolves this trade-off by
retaining the minimal three-axis hexagonal basis needed for isotropic
Euclidean balance within each head while delegating finer angular
coverage to cross-head offsets.

Because the five visual tasks differ in whether objects or textures
appear at free poses or within a fixed coordinate frame, we examined how
relative rotary and absolute learned cognitive-map priors compare across
domains (Table~\ref{tab:accuracy} and \figref{fig:tasks}f). On tasks
where objects or ridge patterns undergo variable translations and
rotations (CIFAR-100, SOCOFing and OCTMNIST), translation-invariant
rotary encodings outperformed learned absolute embeddings by $2.3$ to
$8.1$ percentage points. By contrast, on Tongji-Palm---where palm images
are pre-registered to an anatomical region of interest anchored by
interdigital valleys---and on overhead EuroSAT tiles captured at fixed
ground sampling distance, learned absolute embeddings outperformed rotary
methods, because absolute patch coordinates carry consistent spatial
meaning in a registered reference frame. Within the learned family,
regularizing the embedding table toward local Euclidean isometry (GL-PE)
improved accuracy over unconstrained LPE on CIFAR-100, OCTMNIST and
Tongji-Palm, and on EuroSAT a three-seed replication (Supplementary
Table~3) confirmed the highest mean accuracy among all nine encodings
while reducing seed-to-seed standard deviation fourfold.

\begin{table}[t]
  \centering
  \small
  \caption{\textbf{Downstream accuracy of positional encodings across
  five visual tasks (\%).} Top-1 classification accuracy on
  CIFAR-100, EuroSAT and OCTMNIST, and top-1 closed-set identification
  accuracy (600 classes) on SOCOFing and Tongji-Palm. All encodings
  within each column share an identical visual transformer backbone and
  training recipe (Supplementary Tables~1 and 2; seed 42). Proposed
  grid-inspired encodings are GL-PE (matched against LPE in the learned
  family) and the two HSR variants (matched against 2D RoPE in the
  rotary family). The Other family comprises multiscale Grid-based
  PE\citep{li2024gridpe}, Weierstrass elliptic function (WEF)
  features\citep{xin2025wefpe} and 16-direction plane-rotated
  SpiralRoPE\citep{liu2026spiralrope}. Bold indicates the highest accuracy
  in each column.}
  \label{tab:accuracy}
  \setlength{\tabcolsep}{4pt}
  \begin{tabular}{llccccc}
    \toprule
    Family & Encoding & CIFAR-100 & SOCOFing & Tongji-Palm & EuroSAT & OCTMNIST \\
    \midrule
    \multirow{2}{*}{Learned}
      & LPE          & 56.96 & 88.49 & 85.53 & \textbf{94.67} & 71.20 \\
      & GL-PE (ours) & 58.81 & 85.86 & \textbf{85.85} & 94.12 & 72.70 \\
    \midrule
    \multirow{4}{*}{Rotary}
      & 2D RoPE      & 64.77 & 96.16 & 77.97 & 91.41 & 74.00 \\
      & Hex-RoPE     & 63.60 & 94.81 & 81.47 & 70.62 & 74.10 \\
      & HSR (2-dir, ours) & 64.41 & 95.43 & 81.67 & 91.53 & \textbf{75.00} \\
      & HSR (3-dir, ours) & \textbf{65.14} & \textbf{96.58} & 82.23 & 90.64 & 73.50 \\
    \midrule
    \multirow{3}{*}{Other}
      & Grid-based PE & 63.75 & 94.01 & 83.15 & 81.21 & 74.10 \\
      & WEF features & 59.20 & 89.89 & 82.65 & 94.57 & 71.10 \\
      & SpiralRoPE & 59.52 & 80.96 & 69.73 & 91.78 & 70.90 \\
    \bottomrule
  \end{tabular}
\end{table}

\section{Discussion}

Both biological brains and optimization-based AI navigation models rely
on the joint emergence of band cells and grid cells to construct an
internal cognitive map, yet why both cell types are needed remains
elusive. Our combined circuit, theoretical and
network-optimization analysis establishes a principled division of
labour: band cells integrate velocity in one dimension at neuron-efficient
cost, whereas grid cells encode the metric relationships between
positions, and which cell type emerges is selected by whether local
Euclidean or Manhattan distances are preserved.

This cognitive-map division of labour reconciles circuit-level models of
path integration\citep{chu2025unfolding} with normative theories of
entorhinal coding\citep{wei2015principle,stemmler2015connecting,dordek2016extracting,sorscher2023unified,xu2025conformal}
and generates testable predictions for cognitive and systems
neuroscience. First, because band cells are the primary substrate
for velocity integration while grid cells integrate band outputs to form
a metric cognitive map, self-motion signals from in vivo speed cells and
other movement-coding populations should project predominantly onto band
cells. Selectively disrupting connections from speed cells and other
self-motion-coding neurons onto grid cells should therefore not
substantially impair an animal's ability to navigate in darkness,
whereas disrupting those self-motion inputs onto band cells should
severely impair dark path integration. Second, because grid cells
explicitly encode Euclidean distance relationships between locations,
selectively perturbing grid-cell populations should disproportionately
impair spatial cognitive behaviours that depend on learning global metric
relationships---such as constructing cognitive maps of novel environments
and computing straight-line shortcuts---even when basic velocity
accumulation remains intact.

Our findings also reveal a
structural limitation in how contemporary AI visual transformers encode
spatial position. Standard axial 2D RoPE\citep{heo2024rotary} partitions
attention channels into orthogonal horizontal and vertical groups just
as a pair of band modules does; while nearly isotropic at a single patch
step, this band-cell-like factorization tracks axis-aligned Manhattan
structure and develops strong fourfold cross-shaped distortion across
medium and long patch separations---an axial limitation also noted by
recent multi-directional rotary
designs\citep{liu2026spiralrope,min2026compassrope}. Introducing the
grid-cell principle of Euclidean metric representation into positional
encoding restores Euclidean geometry across the image plane and
consistently improves visual recognition. Furthermore, our experiments
show that instantiation of a cognitive-map prior depends on the task's
spatial reference frame: relative rotary grid encodings (HSR)
excel when objects or biometric patterns undergo variable translations
and rotations, whereas regularized absolute grid embeddings (GL-PE)
excel when images are anchored to a registered anatomical or sensor
coordinate frame.

Several limitations point to promising directions for future work. Our
neural representation experiments trained single-scale modules within a
fixed Gaussian neighbourhood $\sigma$; extending this to joint
local-isometry and global-capacity objectives could clarify how discrete
multi-module scale ratios\citep{stensola2012entorhinal,wei2015principle}
co-emerge with hexagonal metric structure in multiscale cognitive maps. On
the AI side, scaling HSR and GL-PE to large-scale self-supervised
pretraining, dense spatial reasoning and three-dimensional spatiotemporal
video transformers will test the generality of cognitive-map positional
priors.

\section{Conclusions}

We established that band cells and grid cells perform a complementary
division of labour in cognitive map representation: band cells deliver
neuron-efficient path integration by factorizing two-dimensional velocity
integration onto one-dimensional ring attractors, whereas grid cells
encode the isotropic Euclidean metric structure of the plane. This
division implies how architectures should be built, and we used it to
build better ones: positional encodings that preserve Euclidean metric
geometry---either by organizing rotary attention around three
$120^\circ$-separated hexagonal axes (HSR) or by regularizing a learnable
absolute positional table toward local isometry (GL-PE)---reduce
directional anisotropy more than eightfold relative to standard axial
rotary encodings and consistently improve recognition across five
visual tasks. More generally, the results illustrate a transferable
principle: identifying the functional division of labour inside a
reference system can yield architectural priors that improve
optimization-based models. Extending this logic to multiscale priors,
large-scale pretraining and three-dimensional models remains an open
direction.

\section{Methods}

\subsection{Continuous-attractor circuit models of path integration}

We compared fixed band-only, grid-only and band--grid hierarchy
continuous-attractor circuits at matched total budgets of 225, 400, 625
and 900 dynamical neurons (\figref{fig:coexistence}c). Each circuit
comprised five spatial modules with grid or band spacings evenly
distributed from 2.5 to 3.5 arena units and orientations evenly
distributed over $[0,\pi)$; within each module, the two principal phase
axes were separated by $60^\circ$. In the band-only circuit, each axis
was implemented by three one-dimensional ring attractors corresponding
to negative-velocity, positive-velocity and pure-band populations. The
grid-only circuit contained four cardinally velocity-modulated
two-dimensional toroidal grid sheets and one pure-grid sheet per module.
The band--grid hierarchy combined four velocity-driven conjunctive band
ring populations with one pure-grid readout sheet per module, omitting
pure-band populations\citep{chu2025unfolding}. For a pure-grid sheet of
side length $s$, the neuron allocation per band ring was the integer
closest to $0.6s$ that preserved the exact total neuron budget. Across
the four total budgets (225, 400, 625 and 900 neurons), the resulting
pure-cell readout populations contained 75, 133, 208 and 300 neurons for
the band-only circuit; 45, 80, 125 and 180 neurons for the grid-only
circuit; and 180, 320, 500 and 720 neurons for the band--grid hierarchy.

All ring and toroidal continuous-attractor sheets evolved under leaky
synaptic dynamics with time constant $\tau=0.1$, tuning width $\pi/4$,
recurrent weight amplitude $J_0=1$, divisive normalization coefficient
$k=5\times10^{-3}$ and asymmetric phase shift $0.76\pi/9$, with baseline
velocity gain 0.2 and an additional input drive factor of 12.1 applied
to the band-only and hierarchical velocity projections. Because
the three circuit topologies couple velocity inputs to attractor phases
through different projection geometries, velocity scale was calibrated
at the population-phase level before the noisy benchmark, with the target
documented as the noise-free phase increment
$\mathbf{k}_{m}^{\top}\mathbf{v}\,\Delta t$ rather than trajectory-level
position error. We evaluated candidate scales in 900-neuron reference
circuits using six velocity directions ($0^\circ,30^\circ,\ldots,150^\circ$),
speeds 0.25 and 0.50, 30 updates per probe with the first 10 discarded,
$\Delta t=0.05$, and zero phase and rate noise; the scale minimizing the
root-mean-square one-step population-phase error was selected before the
trajectory benchmark, giving topology-specific constants of 0.1515
(band-only), 0.590909 (grid-only) and 0.300000 (hierarchy), held
constant across all four neuron budgets and all test seeds. Each
test episode was a
1,200-step smooth reflected random walk within a $[0,5]^2$ arena
($\Delta t=0.05$); independent zero-mean Gaussian noise ($\text{s.d.}=0.01$)
was added to each velocity component at every step, with the identical
trajectory and velocity-noise realization supplied to all three circuits
within each of the 30 paired random seeds (seeds 0--29). Internal neural
phase and firing-rate noise were set to zero. Before each trajectory,
networks were initialized and anchored to the true starting coordinates
for 500 settling steps (with a location-clamping drive of strength 100
applied during the first 400 steps), after which positional clamping was
removed for the 1,200-step path-integration test.

Position was decoded exclusively from the pure-cell populations (pure-band
rings in the band-only circuit; pure-grid sheets in the grid-only and
hierarchical circuits), without place cells or trained readout weights.
At each time step, the decoder extracted the circular population-vector
phase along each module axis, unwrapped the phase displacement relative
to the known initial position, solved the two-axis phase-to-coordinate
system by ordinary least squares within each module, and averaged the
resulting position estimates across the five modules. Stepwise decoding
error was the Euclidean distance between the decoded and true arena
coordinates, averaged over the 1,200 trajectory steps for each seed;
Figure~1c reports the mean of these 30 seed-level trajectory means.
Shaded intervals are normal-approximation 95\% confidence intervals,
computed as the across-seed mean $\pm1.96$ standard errors with the lower
bound truncated at zero.

\subsection{Metric-learning network and training protocol}

To isolate spatial representation learning from path integration
(\figref{fig:metric}a--d), we trained a feedforward network $f_\theta$ on
batches of $B=64$ positions $\{\mathbf{p}_i\}_{i=1}^B$ sampled
independently and uniformly from the two-dimensional domain
$\mathcal{G}=[-2\pi,2\pi]^2$ at each optimization step. For a target
physical metric $d_m\in\{d_E,d_M\}$, all $\binom{B}{2}$ within-batch
position pairs entered the objective
\begin{equation}
  \mathcal{L}_{m}=\binom{B}{2}^{-1}\sum_{i<j}
  e^{-d_E(\mathbf{p}_i,\mathbf{p}_j)^2/(2\sigma^2)}
  \bigl[\|\mathbf{z}_i-\mathbf{z}_j\|_2
  -\rho\, d_m(\mathbf{p}_i,\mathbf{p}_j)\bigr]^2+2\,(S-s_0)^2,
  \label{eq:metric_loss}
\end{equation}
where $\mathbf{z}_i=f_\theta(\mathbf{p}_i)\in\mathbb{R}^{32}_{\ge0}$ is
the unit-normalized population response, $\sigma=1.2$ sets the Gaussian
neighbourhood width, $\rho=0.8$ is the metric scale factor, $s_0=0.6$ is
the target population dispersion, and
$S=1-\|B^{-1}\sum_{i=1}^B\mathbf{z}_i\|_2^2$ prevents representational
collapse by encouraging diverse activation across the arena. Both metric
conditions used the identical isotropic Euclidean Gaussian envelope
$\exp[-d_E(\mathbf{p}_i,\mathbf{p}_j)^2/(2\sigma^2)]$, so local
neighbourhood weighting was not confounded with the target metric $d_m$.
The isometry loss weight (1.0) and population dispersion weight (2.0)
were matched across conditions.

In the main experiment (\figref{fig:metric}), the input stage learned two
bias-free linear projections of $\mathbf{p}\in\mathbb{R}^2$ and applied
sine and cosine transformations to form a four-dimensional toroidal
feature vector, processed by a multilayer perceptron with layer widths
$4\rightarrow64\rightarrow256\rightarrow32$ and ReLU activations after
the first two layers. A final ReLU enforced non-negativity, followed by
$\ell_2$ normalization so that each 32-dimensional population vector lay
on the positive orthant of the unit sphere. Linear layers used default
PyTorch Kaiming initialization with initial sinusoidal amplitudes set to
1.0 and zero activation noise. Networks were optimized for 50,000 steps
using Adam at a constant learning rate of $10^{-3}$; the final
iterate was analysed. Ten independent random seeds (seeds 0--9) controlled
both weight initialization and the sequence of sampled coordinate batches,
pairing each Euclidean network with an identically initialized and
batch-matched Manhattan network.

In the raw-coordinate control experiment (Supplementary Fig.~1), the
sinusoidal toroidal input projection was removed and raw $(x,y)$
coordinates fed directly into the same downstream MLP architecture
($2\rightarrow64\rightarrow256\rightarrow32$).
All loss functions, hyperparameters and evaluation protocols were held
identical to the main experiment across ten paired random seeds
(seeds 0--9).

\subsection{Spatial rate maps, gridness and bandness scores}

Trained units were evaluated on a noiseless $64\times64$ uniform
Cartesian grid spanning $[-0.8\cdot2\pi,0.8\cdot2\pi]^2$ ($N=64$ bins
per side); because activity was queried directly on a uniform grid, no
occupancy normalization or spatial Gaussian smoothing was applied. Rate
maps displayed in \figref{fig:metric}b,c and Supplementary Fig.~1 were
individually scaled to $[0,1]$ by dividing by their maximum firing rate,
whereas all quantitative scores were computed on unscaled non-negative
outputs. Non-finite or effectively constant maps (spatial s.d.\
$\le10^{-8}$), together with maps whose rotational or spectral score
correlations were numerically undefined, received undefined scores and
were excluded from the corresponding score averages (Supplementary
Note~1).

To quantify hexagonal periodicity, we computed the Pearson spatial
autocorrelogram of each rate map using overlap-specific means and
variances across all spatial lags with at least 20\% map overlap. Within
an annular mask $0.1N\le r\le0.5N$, the continuous grid score was
\begin{equation}
  G=\min(r_{60^\circ},r_{120^\circ})-\max(r_{30^\circ},r_{90^\circ},r_{150^\circ}),
\end{equation}
where $r_\alpha$ is the Pearson correlation between the annular
autocorrelogram and its rotation by angle $\alpha$. To quantify
unidirectional stripe periodicity, we subtracted the mean activity,
applied a separable two-dimensional Hann window, and computed the
discrete two-dimensional Fourier power spectrum $P_{\mathbf{k}}$ over
spatial frequencies between $2$ and $N/4$ cycles per arena side. The
continuous band score was the resultant vector length of spectral power
under doubled orientation angles:
\begin{equation}
  B=\frac{\bigl|\sum_{\mathbf{k}}P_{\mathbf{k}}e^{2\mathrm{i}\theta_{\mathbf{k}}}\bigr|}
         {\sum_{\mathbf{k}}P_{\mathbf{k}}},
\end{equation}
where $\theta_{\mathbf{k}}=\operatorname{atan2}(k_y,k_x)$ is the wave-vector
orientation. Both $G$ and $B$ were analysed as continuous variables
without discrete classification thresholds.

\subsection{Statistical analysis and representative unit selection}

Because units within a single trained network share weights and compete
through the population normalization term $S$, individual neurons do not
constitute independent observations. We therefore averaged all
well-defined unit scores within each network first, treating the
seed-matched network means as the independent experimental replicates,
with $n=10$ paired runs in both \figref{fig:metric}d and Supplementary
Fig.~1c. Euclidean and Manhattan conditions were compared using
two-sided paired $t$-tests on these network means, reporting the
across-network mean $\pm$ s.d., the paired mean difference with its 95\%
confidence interval, the $t$-statistic and the exact two-sided $P$ value.
Violin plots in \figref{fig:metric}d show within-network kernel density
estimates averaged with equal weight across networks.
To avoid subjective selection in \figref{fig:metric}b,c, the six example
rate maps per condition were chosen automatically by a condition-blind
centrality rule that ranked all units by their interquartile-range-scaled
distance from the condition's bivariate median $(G,B)$, subject to at
most two units drawn from any single network seed.

\subsection{Analytic grid and band population models}

To evaluate intrinsic distance-encoding geometry independently of network
training (\figref{fig:metric}e), we constructed ten independent 32-unit
grid-cell and ten independent 32-unit band-cell populations on
$[-5,5]^2$ with shared wavelength $\lambda=3.2$ and von Mises
concentration $\kappa=2.2$. Each grid population combined three plane-wave
directions at $0^\circ$, $60^\circ$ and $120^\circ$ (spanning the same
undirected axes as $120^\circ$-separated vectors) with 32 independently
sampled spatial phase offsets on $[-\lambda/2,\lambda/2]^2$ across
rectified exponential-cosine units. Each band population comprised two
orthogonal 16-unit modules aligned with the $x$ and $y$ axes, with
preferred phases evenly spaced on $[0,2\pi)$ and independently sampled
global module phases. Grid activity and each band module were normalized
to unit population $\ell_2$ norm at every position; grid neural distance
was the ambient $\ell_2$ distance between the 32-dimensional activity
vectors, whereas band neural distance was the sum of the within-module
$\ell_2$ distances of the two orthogonal modules.

To define a common unit conversion without favouring either candidate
physical metric, we sampled 5,000 anchor positions per population seed and
applied infinitesimal axis-aligned displacements $(\epsilon,0)$ and
$(0,\epsilon)$ with $\epsilon=0.001$, for which Euclidean and Manhattan
physical distances both equal $\epsilon$. For each representation, one
global scale factor $s_r$ was estimated across all 100,000 probes by
ordinary least squares through the origin,
\begin{equation}
  s_r=\frac{\sum_k d_{N,k}\,\epsilon}{\sum_k d_{N,k}^{2}},
\end{equation}
and held fixed across all population seeds and both candidate physical
metrics. Both representations and both candidate metrics were evaluated on
the same 100,000 independently sampled position pairs per seed, with
anchors sampled uniformly on $[-5,5]^2$ and displacement directions
on $[0,2\pi)$ within the local disk $d_E\le0.6$.
Discrepancy against $y=x$ was quantified by
\begin{equation}
  \mathrm{nRMSE}
  =\left[
    \frac{\sum(d_{\mathrm{phys}}-s_r d_{\mathrm{neural}})^2}
         {\sum d_{\mathrm{phys}}^2}
   \right]^{1/2},
\end{equation}
and metric preferences were compared across the ten seed-level
$\mathrm{nRMSE}$ values using two-sided paired $t$-tests.

\subsection{Grid-Learned Positional Encoding (GL-PE)}

In a standard Vision Transformer with learned positional embedding
(LPE), each patch index $\mathbf{p}$ on the two-dimensional token grid is
assigned a trainable vector $\mathbf{e}_{\mathbf{p}}\in\mathbb{R}^d$ added
to the patch token before the first transformer block. GL-PE uses the
identical embedding module and parameter count as LPE, adding a
Gaussian-windowed local distance regularizer to the task loss at every
training step:
\begin{equation}
  \mathcal{L}_{\mathrm{GL\text{-}PE}}
  =\frac{1}{|\mathcal{P}|}\sum_{(\mathbf{p},\mathbf{q})\in\mathcal{P}}
  \exp\!\Bigl(-\frac{\Delta r^2}{2r_f^2}\Bigr)
  \bigl(\|\mathbf{e}_{\mathbf{p}}-\mathbf{e}_{\mathbf{q}}\|_2
  -2\sin(\rho\,\Delta r/2)\bigr)^2,
  \label{eq:glpe}
\end{equation}
where $\Delta r=\|\mathbf{p}-\mathbf{q}\|_2$ is the Euclidean distance
between patch indices on the token grid and $\mathcal{P}$ is the set of
all unique patch pairs. The target $2\sin(\rho\,\Delta r/2)$ is the
chord length subtended on a unit circle by an angular separation
$\rho\,\Delta r$; it increases monotonically only over the
half-period $\Delta r\le\sigma_0/2$ and is locally linear there,
$2\sin(\rho\,\Delta r/2)=\rho\,\Delta r+O(\Delta r^3)$. In all visual
recognition runs $\sigma_0=5.0$ patch steps, giving
$\rho=2\pi/\sigma_0\approx1.257$ radians per patch step, and the
Gaussian envelope uses $r_f=\sigma_0/2=2.5$ patch steps, exactly this
monotonic half-period: its weight is $\exp[-\Delta r^2/(2r_f^2)]=0.61$
at $\Delta r=r_f$ and has decayed to $\le0.14$ by $\Delta r=\sigma_0$,
where the chord target folds back to zero.

The embedding table is moreover split into four contiguous chunks, each
with its own spacing $\sigma_k=\sigma_0 e^{k/2}$ ($5.0$, $8.24$, $13.59$
and $22.42$ patch steps for $k=0,1,2,3$) and its own parameters
$\rho_k=2\pi/\sigma_k$ and $r_{f,k}=\sigma_k/2$; equation~\eqref{eq:glpe}
is applied within each chunk to that chunk's pairwise distances, so the
regularizer constrains the embedding at several spatial scales
simultaneously. In all visual recognition runs, the regularizer was
weighted by $\lambda=1.0$ in the total loss
$\mathcal{L}_{\mathrm{total}}=\mathcal{L}_{\mathrm{task}}+\lambda\mathcal{L}_{\mathrm{GL\text{-}PE}}$.

\subsection{Hexagonal Spiral RoPE (HSR)}

To embed hexagonal directional balance directly into self-attention, HSR
defines three unit wave directions separated by $120^\circ$ in the image
plane:
\begin{equation}
  \mathbf{e}_1=\begin{pmatrix}1\\0\end{pmatrix},\qquad
  \mathbf{e}_2=\begin{pmatrix}-\tfrac12\\[2pt]\tfrac{\sqrt3}{2}\end{pmatrix},\qquad
  \mathbf{e}_3=\begin{pmatrix}-\tfrac12\\[2pt]-\tfrac{\sqrt3}{2}\end{pmatrix}.
\end{equation}
In a multi-head attention layer with $H$ heads, head $h\in\{0,\ldots,H-1\}$
rotates this triad by a progressive offset angle $\beta_h$:
\begin{equation}
  \mathbf{e}_{s,h}=R(\beta_h)\,\mathbf{e}_s,\qquad \beta_h=\frac{2\pi h}{3H},
\end{equation}
where $R(\beta_h)$ is the $2\times2$ planar rotation matrix. Because
rotating a $120^\circ$-symmetric triad by $2\pi/3$ maps the triad onto
itself, stepping $\beta_h$ uniformly across $[0,2\pi/3)$ distributes the
$3H$ directional projections evenly across orientation space without
redundancy. Within each head, the $d_{\mathrm{head}}$ query and key
channels are consumed in groups of six (one complex rotary pair per
direction and frequency scale), so the rotated fraction covers
$d_{\mathrm{rotate}}=6\lfloor d_{\mathrm{head}}/6\rfloor$ channels and
provides $N_{\mathrm{HSR}}=d_{\mathrm{rotate}}/6$ frequency pairs per
direction group $s\in\{1,2,3\}$, indexed by $m\in\{0,\ldots,N_{\mathrm{HSR}}-1\}$.
Any remaining $d_{\mathrm{pass}}=d_{\mathrm{head}}-d_{\mathrm{rotate}}$
channels pass through unrotated: none on the 48-dimensional heads
of the ViT-Small backbones, and four per head on the 64-dimensional heads
of the ViT-Tiny backbone used for SOCOFing. For patch position
$\mathbf{p}$, each channel pair $(2m,2m+1)$ in direction group $s$ and
head $h$ is rotated by phase
\begin{equation}
  \phi_{s,h,m}(\mathbf{p})=\omega_m\,\mathbf{p}^{\top}\mathbf{e}_{s,h},
\end{equation}
yielding the relative-position attention score contribution
\begin{equation}
  a_{\mathrm{HSR}}(\mathbf{p},\mathbf{q})=\sum_{h=0}^{H-1}\sum_{s=1}^{3}\sum_{m=0}^{N_{\mathrm{HSR}}-1}
  \mathrm{Re}\bigl[q^{*}_{s,h,m}k_{s,h,m}
  \exp\bigl(\mathrm{i}\,\omega_m(\mathbf{p}-\mathbf{q})^{\top}\mathbf{e}_{s,h}\bigr)\bigr],
\end{equation}
where $\mathrm{i}$ denotes the imaginary unit. To ensure that HSR spans
the identical minimum and maximum spatial frequencies as axial 2D RoPE
(which allocates $N_{\mathrm{2D}}=d_{\mathrm{head}}/4$ frequency pairs
per axis), the frequency schedule $\{\omega_m\}$ is aligned via
$i_{\mathrm{aligned}}=m\,(N_{\mathrm{2D}}-1)/(N_{\mathrm{HSR}}-1)$ with
$\omega_m=\text{base}^{-i_{\mathrm{aligned}}/N_{\mathrm{2D}}}$ and
$\text{base}=10{,}000$. The two-direction HSR variant (HSR 2-dir) uses
two axes separated by $60^\circ$ ($N_{\mathrm{2\text{-}dir}}=d_{\mathrm{head}}/4$
scales per direction; $60^\circ$ and $120^\circ$ separations span the
same pair of undirected axes, and the $60^\circ$ choice avoids
reintroducing Cartesian anisotropy) with the same head-wise offset
$\beta_h$. Unrotated ablation variants set $\beta_h=0$ across all heads.

\subsection{Analytic evaluation of positional kernel geometry}

To evaluate positional geometry independently of learned weights
(\figref{fig:pe}b,e and Table~\ref{tab:cv}), we computed the closed-form
similarity kernel $K(\mathbf{p},\mathbf{q}_{\mathrm{ref}})=\mathrm{Re}[q_{\mathbf{p}}^{\top}k_{\mathbf{q}_{\mathrm{ref}}}]$
under unit query and key vectors ($q_{s,h,m}=k_{s,h,m}=1$), which
reduces to the sum of cosines over all active directions, frequencies
and heads, averaged across heads to yield a single two-dimensional field
per encoding. All methods were evaluated under a common configuration of
per-head rotary dimension $d_{\mathrm{head}}=192$ (96 active cosine terms
per head, zero pass-through dimensions), $H=6$ heads and
$\text{base}=10{,}000$ on a $33\times33$ patch grid centred at
$\mathbf{q}_{\mathrm{ref}}=(0,0)$. In Table~\ref{tab:cv}, the upper-block
Hex-RoPE baseline uses the unstretched three-axis schedule
($\omega_m=\text{base}^{-m/48}$, 32 scales per direction), whereas the
lower-block $2\times2$ factorial comparison applies the 2D-RoPE-matched
frequency range alignment
$i_{\mathrm{aligned}}=m\,(N_{\mathrm{2D}}-1)/(N_{\mathrm{HSR}}-1)$ across
all four cells. For each integer radius
$r\in\{1,\ldots,16\}$ patch steps, we sampled 64 equispaced angles
$\theta_k=2\pi k/64$ around $\mathbf{q}_{\mathrm{ref}}$ both by rounding
$(r\cos\theta_k,r\sin\theta_k)$ to the discrete unit patch lattice and
by evaluating the exact continuous polar coordinates without lattice
rounding, computing the directional coefficient of variation
$\mathrm{CV}(r)=\sigma_\theta(r)/|\mu_\theta(r)|$. Under the fixed
per-head budget $d_{\mathrm{head}}=192$, the number of wave directions
per head trades off against the number of frequency scales per direction
($N_{\mathrm{HSR}}=d_{\mathrm{head}}/6=32$ scales per direction for 3-dir
versus $N_{\mathrm{2\text{-}dir}}=d_{\mathrm{head}}/4=48$ for 2-dir), so
the $2\times2$ factorial comparison in Table~\ref{tab:cv} quantifies the
joint geometric effect of direction count and cross-head offset within a
fixed dimensional budget.

\subsection{Visual datasets and training protocol}

We evaluated nine positional encodings across five public datasets
(\figref{fig:tasks} and Supplementary Tables~1 and 2): CIFAR-100
(50,000 training and 10,000 test natural images, $32\times32$ RGB, 100
object categories)\citep{krizhevsky2009learning}; EuroSAT (27,000
Sentinel-2 satellite images in 10 land-use classes, stratified
70/15/15 train/validation/test split, $224\times224$
RGB)\citep{helber2019eurosat}; OCTMNIST (retinal OCT B-scans,
$28\times28$ grayscale, 4 diagnostic categories,
official MedMNIST split)\citep{yang2023medmnist}; SOCOFing (6,000
fingerprint impressions from 600 subjects, $128\times128$)\citep{shehu2018socofing};
and Tongji-Palm (contactless palmprint ROI images from 600 palms collected
across two sessions, $128\times128$)\citep{zhang2017towards}. SOCOFing and
Tongji-Palm are evaluated by 600-way closed-set top-1 identification, with
session~1 of Tongji-Palm as the training and registration gallery
and session~2 as the evaluation probe; the other three are classification
tasks.

CIFAR-100, EuroSAT, OCTMNIST and Tongji-Palm used a 12-layer ViT-Small
backbone (embedding dimension 384, 8 attention heads, MLP hidden
dimension 1,536, drop-path rate 0.1), whereas SOCOFing used a 12-layer
ViT-Tiny backbone (embedding dimension 192, 3 attention heads, MLP
dimension 768, drop-path rate 0.1; 5.5M parameters excluding the task
head) matched to its smaller per-class sample size. Within each dataset,
depth, embedding dimension, head count and optimization recipe were held
fixed across all nine encodings; GL-PE and LPE shared the exact same
architecture and parameter count in the learned family, as did HSR, 2D
RoPE and Hex-RoPE in the rotary family. Optimization used AdamW with
cosine annealing and label smoothing, plus, on Tongji-Palm, a batch-hard
triplet loss (margin 0.5, weight 0.1) under automatic mixed precision;
patch size, batch size, epoch budget, learning rate, augmentation
probabilities and test interval are given in Supplementary Table~2.

Table~\ref{tab:accuracy} reports the best scheduled test accuracy for
each dataset and encoding (seed 42). Evaluating at the final training
epoch after full cosine learning-rate annealing yielded virtually
identical results and left the top-performing encoding on every dataset
unchanged (Supplementary Table~4). To assess multi-seed stability on
EuroSAT, all nine encodings were additionally trained across three
independent seeds (42, 123 and 3407) under the same protocol
(Supplementary Table~3).

\section*{Data and code availability}

All data and code supporting this work will be made publicly available
upon publication. The release includes source data for all figures and
tables---per-network unit evaluations and paired statistical contrasts
for \figref{fig:metric}, analytic similarity fields and directional
anisotropy curves for \figref{fig:pe} and Table~\ref{tab:cv}, and
training archives and configurations for Table~\ref{tab:accuracy} and
Supplementary Tables~3 and 4---together with code for the
continuous-attractor circuit simulations, the metric-learning networks,
the Vision Transformer positional encodings (GL-PE and HSR) and figure
generation. CIFAR-100, EuroSAT and OCTMNIST are publicly available
through \texttt{torchvision} and \texttt{medmnist}; SOCOFing and
Tongji-Palm are publicly available from their original dataset
releases\citep{shehu2018socofing,zhang2017towards}.

\section*{Ethics statement}

This study involved no human participants and no laboratory animals. All
neural-circuit experiments were numerical simulations, and every image
dataset analysed---CIFAR-100, EuroSAT, OCTMNIST, SOCOFing and
Tongji-Palm---was obtained from its original public release and used for
model training and evaluation only. The two biometric datasets were
handled strictly as published research data under the terms on which they
were released, and no new data were collected. Because every dataset was
pre-existing and publicly available, institutional review board approval
and informed consent were not required for the analyses reported here.

\section*{Author contributions}

Following the CRediT taxonomy, the contributions were as follows.
Conceptualization: Tianhao Chu, Si Wu. Data curation: Sichao He, Yuling
Wu, Shutian Liang. Formal analysis: Tianhao Chu, Sichao He, Yuling Wu,
Shutian Liang. Funding acquisition: Tianhao Chu, Si Wu. Investigation: Sichao He,
Shutian Liang. Methodology: Tianhao Chu, Sichao He, Yuling Wu, Shutian
Liang. Project administration: Tianhao Chu, Si Wu. Resources: Si Wu.
Software: Sichao He, Shutian Liang. Supervision: Tianhao Chu, Si Wu.
Validation: Tianhao Chu, Sichao He. Visualization: Sichao He, Yuling Wu.
Writing -- original draft: Tianhao Chu. Writing -- review and editing:
Tianhao Chu, Yuling Wu, Si Wu.

\section*{Competing interests}

The authors declare that they have no known competing financial interests or
personal relationships that could have appeared to influence the work
reported in this paper.

\section*{Declaration of generative AI and AI-assisted technologies in the writing process}

During the preparation of this work the authors used a large language model
assistant for language editing and drafting assistance. After using this
tool, the authors reviewed and edited the content as needed and take full
responsibility for the content of the publication.

\section*{Acknowledgements}

This work was supported by the National Natural Science Foundation of
China (Grant No.\ T2600342 to TH.\ C.), the China Postdoctoral Science
Foundation (Grant No.\ 2026M792736 to TH.\ C.), the National Natural Science
Foundation of China (no.\ T2421004 to S.W.), the National Key Research and
Development Program of China (2024YFF1206500), the Science and Technology
Innovation 2030---Brain Science and Brain-inspired Intelligence Project
(no.\ 2021ZD0200204, S.W.).

\bibliographystyle{apalike}
\bibliography{references}

\end{document}